\documentclass{article} 
\usepackage{colm2024_conference}

\usepackage{microtype}
\usepackage{hyperref}
\usepackage{url}
\usepackage{booktabs}
\definecolor{darkblue}{rgb}{0, 0, 0.5}
\hypersetup{colorlinks=true, citecolor=darkblue, linkcolor=darkblue, urlcolor=darkblue}

\usepackage{times}
\usepackage{latexsym}
\usepackage{algorithm}
\usepackage{algorithmic}
\usepackage[utf8]{inputenc} 
\usepackage[T1]{fontenc}    
\usepackage{url}            
\usepackage{booktabs}       
\usepackage{amsfonts}       
\usepackage{nicefrac}       
\usepackage{microtype}      
\usepackage{xcolor}         
\usepackage{graphicx}
\usepackage{caption}
\usepackage{subfigure}
\usepackage{multirow}
\usepackage{arydshln}
\usepackage{booktabs}
\usepackage{amsmath}
\title{ArtGPT-4: Towards Artistic-understanding \\ Large Vision-Language Models with Enhanced Adapter}

\author{Zhengqing Yuan$^{1}$, Yunhong He$^{3}$, Kun Wang$^{3}$\footnotemark[1], Yanfang Ye$^{1}$, Lichao Sun$^{2}$\footnotemark[1] \\
$^{1}$University of Notre Dame\quad
$^{2}$Lehigh University\quad
$^{3}$Anhui Polytechnic University\\
}
\colmfinalcopy
\begin{document}

\maketitle
\renewcommand{\thefootnote}{\fnsymbol{footnote}}
\footnotetext[1]{Kun Wang and Lichao Sun are co-corresponding authors: \href{mailto:kun.wang@ahpu.edu.cn}{\color{black}{kun.wang@ahpu.edu.cn}} and \href{mailto:lis221@lehigh.edu}{\color{black}{lis221@lehigh.edu}}.}

\begin{abstract}
The success of large language models (LLMs) has inspired an emerging research field of multimodal learning. However, a grand challenge of exploiting LLMs for multimodal learning is the size of pre-trained LLMs which are always with billions of parameters. To tackle this challenge, models such as MiniGPT-4 and LLaVA have been developed to fine-tune the pre-trained models using fewer parameters. Despite their promising performance, these models remain limited in their understanding of artistic imagery. To facilitate better artistic-understanding, in this paper, we propose ArtGPT-4, a pioneering large vision-language model tailored to address the limitations of existing models in artistic comprehension. The key innovation of ArtGPT-4 lies in its craft for the sophisticated challenge of artistic image comprehension, setting it apart from other models that overlook fine details for broader themes. Specifically, it works by integrating some specialized adapter layers into the LLM, enabling the model to more efficiently and effectively parse and interpret complex visual tokens, instead of fine-tuning the whole LLM as in the existing method. ArtGPT-4 has demonstrated its outstanding performance on the efficiency: utilizing a Tesla A100 device, its training can be completed in mere 2 hours with an image-text pair dataset comprising approximately 0.52M entries. Additionally, ArtGPT-4 has also achieved state-of-the-art performance on the ArtEmis and ArtEmis-v2.0 datasets as well as the benchmarks established in this work, lagging behind professional artists' descriptions by a negligible 0.15 points on a 6-point scale. The outstanding performance of ArtGPT-4 shows that it can render images with an artistic-understanding and convey the emotions they inspire, mirroring human interpretation. The code and the pre-trained model are accessible in \url{https://github.com/DLYuanGod/ArtGPT-4}.
\end{abstract}

\section{Introduction}

Advancements in large language models (LLMs) have revolutionized the field of natural language processing, paving the way for numerous breakthrough applications and sophisticated tasks~\citep{ouyang2021training,brown2020language,openai2022chatgpt}. On the other hand, single-modality LLMs, as powerful as they are, represent just one facet of the broader potential of artificial intelligence (AI). The budding realm of multimodal models, which synergize different data modalities like text and vision, holds promise for a new wave of innovations. Notable works in this direction include recent studies by Lin et al.~\citep{lin2021m6} and Radford et al.~\citep{radford2021learning}, indicating the infancy but undeniable potential of this domain. GPT-4, a monumental achievement by OpenAI, has recently eatablished a benchmark in the vision-language understanding sphere~\citep{openai2023gpt4}. Its prowess in discerning intricate visual nuances and producing varied, contextually rich language outputs is nothing short of groundbreaking. However, the lack of open-source availability for GPT-4 poses challenges for the broader research community. Without access to its architecture, replicating or building upon its success becomes a convoluted endeavor. Furthermore, the sheer volume of data that GPT-4 relies on - i.e., amassing over 45 terabytes of text or image information - raises questions about the feasibility of gathering comparable datasets for similar projects. Given these constraints, leveraging existing pre-trained models and fine-tuning them for specific tasks presents a pragmatic and increasingly popular strategy for researchers and practitioners.

\begin{figure}[t]
\centering
  \includegraphics[width=0.73\textwidth]{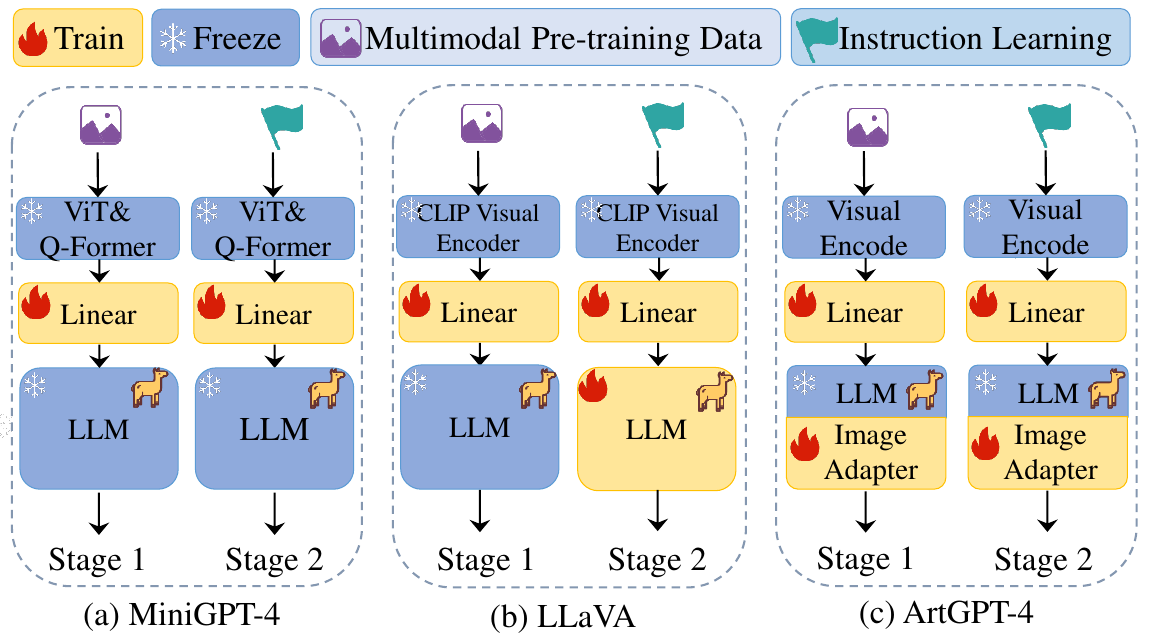}
  \caption{Comparison between different structures of multimodal models. All of these methods are trained in a two-stage fashion. Stage 1 stands for pre-training and Stage 2 represents instruction tuning.}
  \label{fig:4}
  \vspace{-10pt}
\end{figure}


\begin{figure*}[h]
  \includegraphics[width=\textwidth]{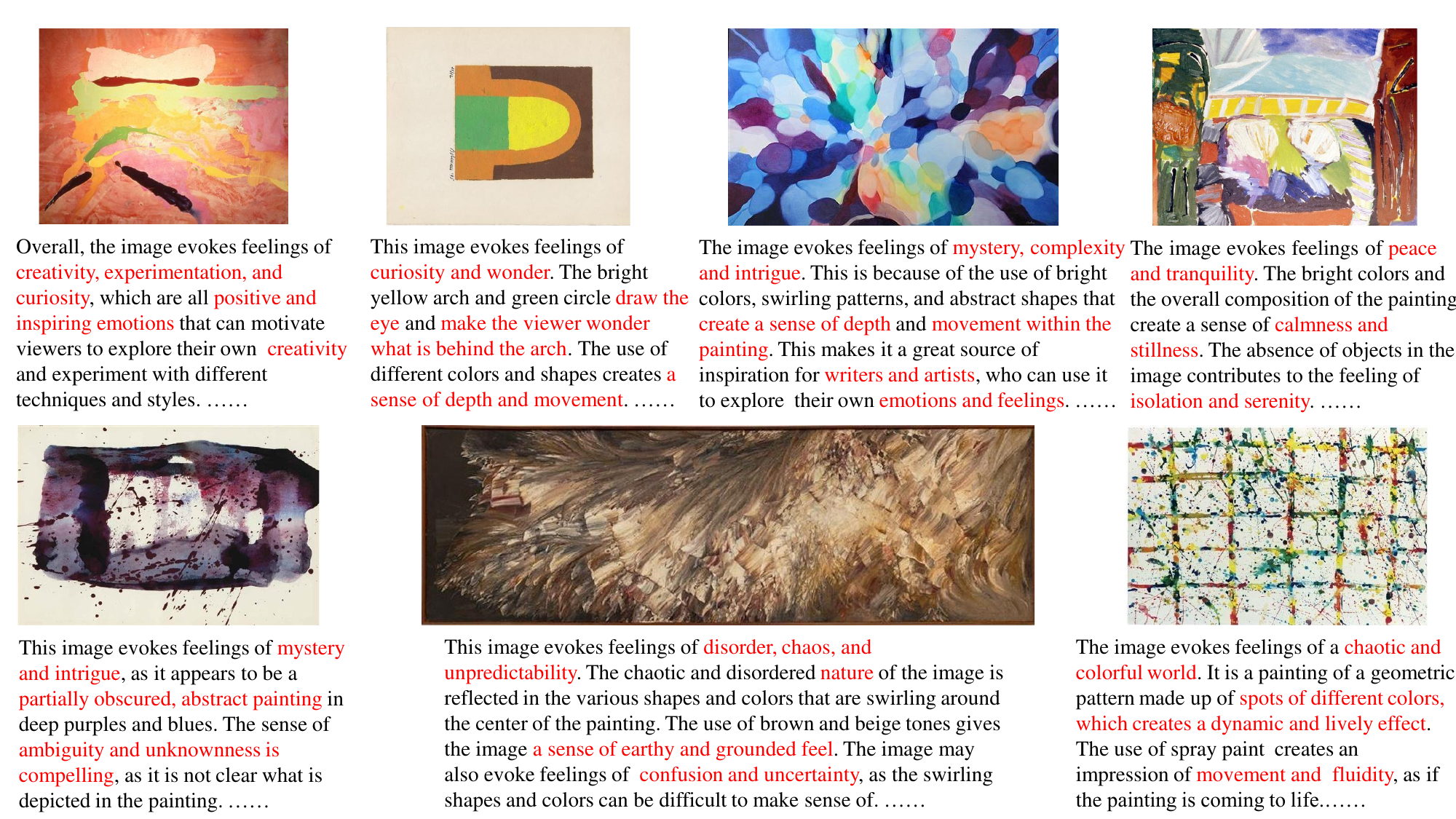}
  \caption{ArtGPT-4 exhibits a remarkable ability of artistic-understanding. It extends beyond merely capturing the artistic details of an image, delving into the realm of emotional understanding. ArtGPT-4 is capable of discerning and articulating the emotions elicited by an image like a human being, such as feelings of positivity and inspiration.}
  \label{fig:teaser}
  
\end{figure*} 



As shown in Figure~\ref{fig:4}, there have been two primary open source models of end-to-end multimodal LLMs: (1) MiniGPT4~\citep{zhu2023minigpt} model employs a limited parameter set and maintains frozen LLM and visual models during both pretraining and instruction tuning phases. A major limitation of that this model is its reliance on static visual models and LLMs, which may result in suboptimal alignment due to the constrained parameter count. (2) LLaVA~\citep{liu2023visual} model incorporates trainable LLMs during instruction tuning while keeping the visual models static. However, a significant challenge with this model is the computational cost, as updating all LLM parameters during training can be resource-intensive. Although these models have demonstrated impressive performance across a broad spectrum of tasks, including image understanding and detail depiction, matching the proficiency of GPT-4, they leave a large room for improvement when tasked with the nuanced interpretation of artistic images akin to human perception. Specifically, current multimodal models fall short in capturing the intricate details inherent in an art image and articulating the emotions it elicits in an objective manner like a human observer.



In this paper, we propose ArtGPT-4, a novel model designed to address the aforementioned limitations of existing multimodal models for artistic-understanding. Specifically, ArtGPT-4 incorporates tailored linear layers and their corresponding activation functions exclusively into the language model, in tandem with the activation of specific training parameters. These modifications are strategically implemented to optimize the model's performance and equip it to effectively tackle the challenges of artistic understanding inherent in vision-language tasks. This approach distinctly differentiates it from existing models, which often finetune the whole LLM to capture more detail about visual features like LLaVA series models. Trained on a Tesla A100 device in a mere span of 2 hours, ArtGPT-4 utilized only 0.52M image-text pairs, amounting to about 200GB. The model can depict images with an enhanced artistic flair and convey the emotions they inspire, as shown in Figure~\ref{fig:teaser}. Subsequent evaluation methods revealed that ArtGPT-4 outperforms existing models in the realm of artistic image understanding. Our contributions are as follows:

\begin{itemize}
    \item In this work, we pioneer the exploration of artistic understanding within multimodal models. It addresses the inherent limitations of existing multimodal models, which have struggled to comprehensively grasp the intricate nuances of artistic imagery and to objectively articulate the emotions they elicit in a manner reminiscent of human perception.
    \item Our proposed ArtGPT-4 introduces a parameter-efficient fine-tuning method exclusively for the language components of multimodal models at the first attempt. It has yielded remarkable outcomes, effectively addressing the challenges associated with the extensive time and resource demands of training large visual-language models.
    \item We have also established a novel dataset tailored for assessing the visual comprehension capabilities of multimodal models, which holds potential for a more in-depth evaluation of expansive vision-language models.
\end{itemize}

\section{Related Work}


\noindent\textbf{Vision-Language Model.}~~ In recent years, the pursuit of models with capabilities transcending a single domain has gained momentum. A notable exemplar in this realm is OpenAI's CLIP~\citep{radford2021learning}, which pioneered the synergy between visual and linguistic understanding by associatively training on image-text pairs. Building on such foundational work, researchers have further delved into models that empower language architectures with image comprehension capabilities~\citep{chen2022visualgpt,alayrac2022flamingo,tsimpoukelli2021multimodal}. Innovative training methodologies have emerged for these multimodal models. A case in point is BLIP-2~\citep{li2023blip}, which introduces a versatile and efficient pre-training paradigm for vision-language endeavors. This approach capitalizes on readily available frozen pre-trained image encoders and expansive language models, complemented by a nimble Q-Former for mapping modules to bridge the modality chasm, as depicted in Figure~\ref{fig:teaser} (b). Notably, MiniGPT-4, utilizing the BLIP-2 architecture, harnesses the capabilities of a pre-trained ViT, Q-Former, and integrates with the Vicuna model to achieve profound image understanding proficiencies.

\noindent\textbf{Efficient Fine-tuning.}~~ Parameter-efficient fine-tuning techniques~\citep{houlsby2019parameter,zaken2022bitfit,li2021prefix,he2021towards,qing2023mar} have gained traction in the NLP domain. These methods aim to minimize the number of learning parameters and computational resources needed for downstream task adaptation, yet they achieve results comparable to full fine-tuning. In the realm of computer vision, there has been a surge in research focused on efficient learning. Works by Jia et al.~\citep{jia2022visual}, Bahng et al.~\citep{bahng2022visual}, and Chen et al.~\citep{chen2022adaptformer} have delved into visual adaptation using methodologies akin to those in NLP. However, it's pivotal to note that these studies primarily focus on adaptations within the same modality such as text-to-text, image-to-image, video-to-video, or within the same domain~\citep{yang2023aim} like image-to-video.

\begin{figure*}[t]
    \centering
  \includegraphics[width=.9\textwidth]{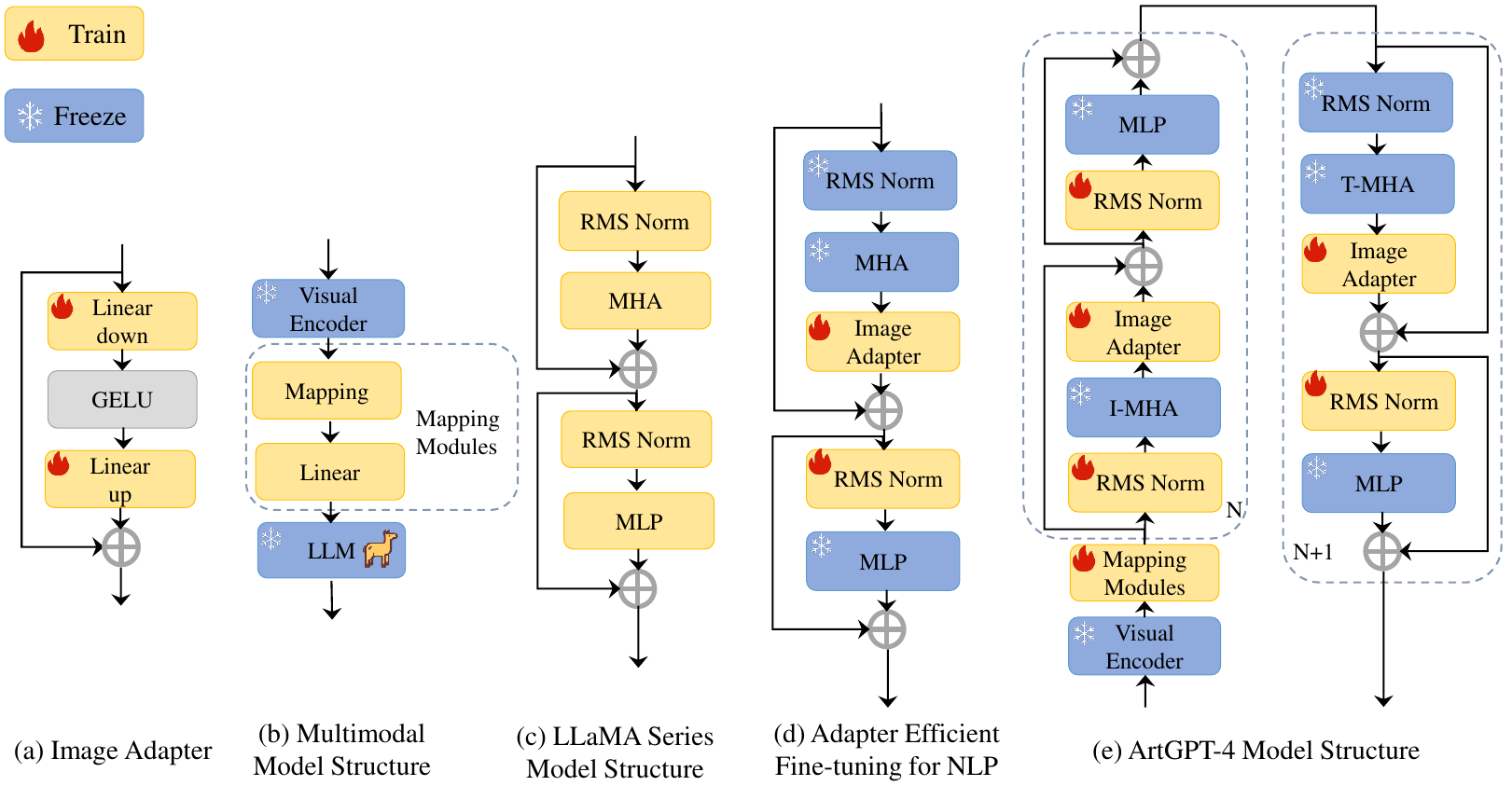}
  \caption{We show how we adapt the LLM (c) of Multimodal Model Structure (b) using the Adapter Efficient Fine-tuning method in NLP to model the ArtGPT-4 (e). During training, only newly added Image Adapters (a) and partial normalization layer (e) are updated while all the other layers are frozen.}
  \vspace{-5pt}
  \label{fig:3}
\end{figure*}

\section{ArtGPT-4}
In this section, we will detail the structure of ArtGPT-4, and the training steps of ArtGPT-4, illustrating how we use the enhanced Adapter layer to construct visual-language multimodal models.

\subsection{Image Adapter}
Drawing inspiration from the advancements in finetuning techniques within the realms of Natural Language Processing (NLP) and Computer Vision (CV)~\citep{houlsby2019parameter,zaken2022bitfit,li2021prefix,he2021towards,qing2023mar,yang2023aim}, we have integrated the Adapter mechanism, as proposed by Houlsby et al.\citep{houlsby2019parameter}. Figure~\ref{fig:3}(a) depicts the Adapter's architecture, characterized by its bottleneck design. This design consists of two linear layers separated by an activation layer. The primary function of the initial linear layer is to diminish the input's dimensionality, whereas the subsequent layer restores it to its original dimension, as shown in Equation~\ref{adapter}.

\begin{align} \label{adapter}
\mathbf{Y}_{adp} &= \text{Adapter}(\mathbf{Y}_{MHA};\mathbf{W}_{down},\mathbf{W}_{up}) \nonumber \\
&= \mathbf{W}_{up} \left( \text{GELU} \left( \mathbf{W}_{down} \mathbf{Y}_{MHA} \right) \right) + \mathbf{Y}_{MHA}
\end{align}
where $\mathbf{Y}_{MHA}$ represents the data computed from the multi-head attention layer (MHA). The symbols $\mathbf{W}_{down}$ and $\mathbf{W}_{up}$ respectively signify the trainable weight matrices responsible for input dimension reduction and its subsequent restoration.

To fine-tune pre-existing models for downstream NLP tasks more efficiently, we positioned an Adapter subsequent to the MHA layer, as shown in Figure~\ref{fig:3}(c) and (d). To further ensure training stability after the integration of the Adapter, we will update the parameters of the normalization layer, specifically the RMS Norm. The computational representation of this block is

\begin{align} \label{AFRMS}
\mathbf{Y}_{RMS} &= \text{RMS Norm}_{2}\left ( \mathbf{Y}_{adp};\mathbf{W}_{RMS} \right ) \nonumber \\
& = \mathbf{W}_{RMS}\odot \frac{\mathbf{Y}_{adp}}{\sqrt{\text{mean}\left ( \mathbf{Y}_{adp}^{2} \right ) + \mathbf{\epsilon }} } 
\end{align}
where $\mathbf{Y}_{RMS}$ is the resultant data post-RMS Norm normalization, and $\mathbf{W}_{RMS}$ is the learnable weight matrix. The symbol $\odot$ denotes elemental multiplication, and $\epsilon$ is a minuscule constant introduced to prevent zero denominators.

As shown in Figure~\ref{fig:3}(b), to capture image and instruction information, mapping modules are usually incorporated prior to pre-trained language models because it is commonly believed that language models can embed visually structured information from tokens~\citep{radford2021learning,li2022blip}. As shown in the following formula

\begin{align}
&\mathbf{Y}_{v}  = \text{Visual Encode} \left ( \mathbf{X}_{v} \right ) \nonumber  \\
&\mathbf{X}_{image-token} = \mathbf{W}_{Map} \mathbf{Y}_{v} \nonumber \\
&\mathbf{X} = \left [ \mathbf{X}_{image-token},\mathbf{X}_{text-token} \right ] 
\end{align}
where $\mathbf{X}_{v}$ denotes the image information, Visual Encode$ (\cdot)$ denotes the visual encoder model, $\mathbf{W}_{Map}$ denotes the weight matrix of the trainable mapping layer, $\mathbf{X}_{text-token}$ denotes the textual information, and $\mathbf{X}$ denotes the information that will be inputted to the LLMs containing the image and text.

However, the introduction of new mapping modules which need to fully fine-tune LLM can lead to an excessive number of adjustable parameters~\citep{liu2023visual} or result in inadequate alignment when freezing full LLM during training~\citep{zhu2023minigpt}. To tackle these challenges, we propose two novel strategies: 1) repurposing the pre-trained self-attention layer in the language model for visual modeling, and 2) introducing new Image Adapter modules to these pre-trained language models. More specifically, in the first strategy, we denote the original self-attention layer as T-MHA for linguistic modeling, and the reused T-MHA layer as I-MHA for visual modeling. As depicted in Figure~\ref{fig:3}(e), I-MHA precedes T-MHA. The primary differentiation between T-MHA and I-MHA is their data input processing. Specifically,  input data of I-MHA, $\mathbf{Y}_{RMS1}$, employs an RMS Norm with learnable parameters, enhancing the normalization of image-containing tokens and thereby boosting the computational efficiency of I-MHA.

\begin{align} \label{RMS_1}
&\mathbf{Y}_{RMS1} = \text{RMS Norm}_{\text{I1}}\left ( \mathbf{X};\mathbf{W}_{RMS1} \right ) \nonumber \\
&\mathbf{Y}_{I-MHA} = \text{I-MHA}\left ( \mathbf{Y}_{RMS1} \right )
\end{align}
Where $\text{I-MHA}\left ( \cdot \right )$ represents the T-MHA layer, which remains static in terms of parameter updates. $\mathbf{Y}_{I-MHA}$ denotes the output data post I-MHA computation.

\noindent To further enhance the alignment between image and text data for the second strategy, ArtGPT-4 incorporates trainable adapters, termed the Image Adapter, as showcased in Figure~\ref{fig:3}(e). The computation of this block can be written as

\begin{align}
&\mathbf{Y}_{T-MHA} = \text{T-MHA}\left (\text{RMS Norm}_{\text{T1}}\left ( \mathbf{Y} \right ) \right ) \nonumber \\
&\mathbf{Y}_{Iadp}=\text{Image Adapter} \left (\mathbf{Y}_{I-MHA/T-MHA} ;\mathbf{W}_{down},\mathbf{W}_{up} \right ) \nonumber \\
&\mathbf{Y}_{RMS2} =  \text{RMS Norm}_{2} \left ( \mathbf{Y}_{Iadp} ;\mathbf{W}_{RMS2} \right ) 
\end{align}
In this representation, $\text{T-MHA}\left ( \cdot \right )$ signifies the T-MHA layer, and $\text{RMS Norm}_{\text{T1}}\left ( \cdot \right )$ indicates the RMS Norm layer preceding the T-MHA layer. Both these layers remain unaltered in terms of parameter updates. The symbols $\mathbf{Y}$, $\mathbf{Y}_{T-MHA}$, $\mathbf{Y}_{Iadp}$, and $\mathbf{Y}_{RMS2}$ represent the input data, post T-MHA layer data, post Image Adapter output data, and post second RMS Norm layer data, respectively.

\section{Training}\label{Training Process}
  ArtGPT-4 remains to enable Language Models to artistic-understand visual information using pre-trained models. We still follow the parameters of the original Multimodal Model, like MiniGPT-4 or LLaVA, and its training steps. Only we use different training data and model structures based on these original parameters, as shown in Figure~\ref{fig:4}(c).

 \noindent\textbf{Training Data.}~~We use Laion-aesthetic from the LAION-5B~ \citep{schuhmann2022laion} dataset, which amounts to approximately 200GB for the first 0.52M data. The aesthetic of this dataset quality is a scale from 7 to 10, while affective polarity is rated as positive, neutral, or negative. In addition to image ratings, the dataset also includes metadata such as image tags and image descriptions.

\noindent\textbf{The First Stage Training Processes.}~~We trained our model using the following hyperparameters: a linear warmup cosine learning rate scheduler with an initial learning rate of 1e-7, a minimum learning rate of 8e-7, and a warmup learning rate of 1e-8. The weight decay was set to 0.05, and the maximum number of training epochs was 2. We used a batch size of 32 for both training and evaluation, with 4 workers. The warmup steps were set to 5000, and there were 5000 iterations per epoch. We only trained on a Tesla A100 for less than 2 hours using the Laion-aesthetic dataset.

\noindent\textbf{The Second Stage Training Processes.}~~We fine-tuned the ArtGPT-4 using a set of MiniGPT-4 or LLaVA's image-text pairs and instructions, such as ``\textit{<Img><ImageHere></Img> Take a look at this image and describe what you notice.\#\#\#Assistant.}''~~

We employed the same template containing a prompt with a randomly sampled instruction, which allowed our model to generate more natural and reliable responses. Specifically, as shown in Equation~\ref{eq:1_1}:

\begin{equation}\label{eq:1_1}
    p\left ( \mathbf{X}_{a}\mid \mathbf{X}_{v},\mathbf{X}_{instr} \right ) =\prod_{i}^{L} p_{\theta} \left ( x_{i}\mid \mathbf{X}_{v},\mathbf{X}_{a,<i},\mathbf{X}_{instr}  \right )
\end{equation}
where $\mathbf{X}_{instr}$ denotes the instruction randomly selected from Table 1 in the Supplementary Material, $\mathbf{X}_{a}$ denotes the answer to the image by the model for that instruction, $\theta$ denotes the training parameters of the model, and $\mathbf{X}_{a,<i}$ are the instruction and answer tokens from all previous rounds prior to the current prediction token $x_{i}$. We only trained on a Tesla A100 for less than 10 minutes.

\section{Evaluation}

\subsection{Zero-shot Testing Datasets}

We use \textbf{ArtEmis}~\citep{achlioptas2021artemis}, \textbf{ArtEmis-v2.0}~\citep{youssef2022artemis2}, \textbf{ArtMM} as zero-shot testing datasets, a detailed description of these datasets is given in Appendix \ref{Datasets}.



\subsection{Baselines}
\noindent\textbf{MiniGPT-4~\citep{zhu2023minigpt}.}~~It is a streamlined model that merges a visual encoder with the LLM, showcasing multi-modal abilities akin to GPT-4. Through fine-tuning with a quality dataset and conversational approach.

\noindent\textbf{LLaVA~\citep{liu2023visual}.}~~LLaVA is a large multimodal model combining a vision encoder with an LLM, utilizing GPT-4 to generate multimodal instruction-following data. Early tests indicate LLaVA's exceptional chat abilities, rivaling GPT-4V in some areas.

\noindent\textbf{Mulit-modal GPT~\citep{gong2023multimodalgpt}.}~~This is a vision and language model designed for multi-round dialogues with humans, which is fine-tuned from OpenFlamingo with the addition of LoRA in both cross- and self-attention area.

\noindent\textbf{VisualGPT~\citep{chen2022visualgpt}.}~~It is a data-efficient image captioning model that utilizes linguistic knowledge from a LLM. Its solution presented is a unique self-resurrecting encoder-decoder attention mechanism that adapts the LLM with limited in-domain data. 

\noindent\textbf{GIT~\citep{wang2022git}.}~~GIT is designed to merge vision-language tasks like image captioning and question answering. It simplifies the design with just an image encoder and text decoder, focusing on a singular language modeling task.

\noindent\textbf{ViLT~\citep{kim2021vilt}.}~~ ViLT (Vision-and-Language Transformer) model lies in its monolithic nature, wherein the handling of visual data is radically streamlined, eliminating the need for convolution and treating it similarly to textual data.

\section{Evaluation Metrics}
\noindent\textbf{VADER.}~~ For VADER-based similarity~\citep{hutto2014vader}, we utilized the VADER sentiment analyzer to compute the compound sentiment score for each response. The compound sentiment score represents the overall sentiment of the response, ranging from -1 (most negative) to 1 (most positive). The sentiment similarity between two responses, model response and Labeling of data sets, was then calculated as the absolute difference between their respective compound sentiment scores. A higher score indicates better performance.

\noindent\textbf{TextBlob.}~~ For TextBlob-based similarity~\citep{loria2018textblob},, we used the TextBlob library to analyze the polarity (sentiment score) of each response. The polarity ranges from -1 (most negative) to 1 (most positive). Similar to VADER-based similarity, the sentiment was computed as the absolute difference between their polarity scores. Higher scores indicate better performance.

\noindent\textbf{BERT.}~~ For BERT-based similarity, we employed the SentenceTransformer BERT~\citep{reimers2019sentence} model to encode text into high-dimensional embeddings. We then utilized the cosine similarity metric to quantify the similarity between their embeddings. The cosine similarity, as shown in 
Equation (\ref{COS}) ranges from -1 (completely dissimilar) to 1 (identical). Higher scores indicate better performance.

\begin{equation} \label{COS}
    \cos \theta =\frac{  \mathbf{Y}_{A} \cdot \mathbf{Y}_{B}}{\left \| \mathbf{Y}_{A} \right \| \times \left \| \mathbf{Y}_{B} \right \|}
\end{equation}
Where $\mathbf{A}$ denotes the word embedding vector of the model responses and $\mathbf{B}$ denotes the word embedding vector of the labels in the dataset. The $\mathbf{A} \cdot \mathbf{B}$ denotes the dot product of vectors $\mathbf{A}$ and $\mathbf{B}$ and $\left \| \cdot \right \|$ denotes the Euclidean norm.



\begin{table*}[t]
\resizebox{.95\hsize}{!}{
\begin{tabular}{l|cc|ccc|ccc}
\toprule
\hline
\multirow{2}{*}{Methods}             & \multicolumn{1}{l}{Pretraing} & \multicolumn{1}{l|}{Learnable} & \multicolumn{3}{c|}{ArtEmis} & \multicolumn{3}{c}{ArtEmis-v2.0} \\
                                              & \multicolumn{1}{l}{Language model}                & parameters (\textbf{B}illion)            & VADER   & TextBlob  & BERT   & VADER    & TextBlob    & BERT    \\ \hline
MiniGPT-4~\citep{zhu2023minigpt}                                     & Vicuna-13B                    & 0.003B                              & 0.746   & 0.242     & 0.693  & 0.939    & 0.332       & 0.673   \\

\multicolumn{1}{l|}{MiniGPT-4~\citep{zhu2023minigpt}}            & Vicuna-7B                    & 0.003B                            & 0.704          & 0.225          & 0.684          & 0.901          & 0.321          & 0.660   \\

\multicolumn{1}{l|}{MiniGPT-4~\citep{zhu2023minigpt}}            & Alpaca-7B                    & 0.003B                            & 0.660          & 0.211          & 0.666          & 0.853          & 0.319          & 0.649   \\
LLaVA ~\citep{liu2023visual}                                               & Vicuna-13B                    & \underline{13B}                            & 0.740   & 0.240     & 0.688  & 0.935    & 0.332       & 0.673   \\
Multi-modal GPT ~\citep{gong2023multimodalgpt}                               & OpenFlamingo-9B               & 1.66B                          & 0.701   & 0.234     & 0.592  & 0.780    & 0.301       & 0.599   \\
VisualGPT ~\citep{chen2022visualgpt}                                     & GPT-2-small                   & 0.124B                           & 0.105   & 0.098     & 0.122  & 0.141    & 0.101       & 0.155   \\
GIT ~\citep{wang2022git}                                          & -                             & 0.7B                           & 0.101   & 0.072     & 0.110  & 0.155    & 0.100       & 0.139   \\
ViLT  ~\citep{kim2021vilt}                                        & BERT                          & 0.0874B                          & 0.024   & 0.011     & 0.104  & 0.082    & 0.110       & 0.109   \\ 
\hline

ArtGPT-4 (Backbones MiniGPT-4)                                    & Vicuna-7B  &                  0.26B                          & 0.740 & 0.233 & 0.686 & 0.920 & 0.327 & 0.665   \\
ArtGPT-4 (Backbones MiniGPT-4)                                    & Alpaca-7B  &                  0.26B                          & 0.721 & 0.233 & 0.685 & 0.923 & 0.326 & 0.669   \\
ArtGPT-4 (Backbones LLaVA)                                    & Vicuna-13B  &              0.52B                          & 0.799 & 0.245 & 0.691 & 0.982 & 0.350 & 0.689   \\
ArtGPT-4 (Backbones MiniGPT-4)                                    & Vicuna-13B                    & 0.52B                          & \textbf{0.813}   & \textbf{0.247}     & \textbf{0.693}  & \textbf{0.987}    & \textbf{0.360}       & \textbf{0.698}   \\
\hline
\bottomrule
\end{tabular}}
\caption{\label{Reasult} Evaluation on ArtEmis and ArtEmis-v2.0 with six state-of-the-art multi-modal models.}
\end{table*}

\begin{table*}[t]
\centering
\resizebox{.95\hsize}{!}{
\begin{tabular}{l|cccccccc|c}
\toprule
\hline
\multirow{2}{*}{Methods}  & \multicolumn{2}{c}{IDC} & \multicolumn{2}{c}{ISAC} & \multicolumn{2}{c}{ICRC} & \multicolumn{2}{c|}{MDIUC} & \multirow{2}{*}{Total average} \\
   & sum      & average      & sum       & average      & sum       & average      & sum        & average       &                                \\ \hline
Artist (Human)              & 41       & 4.1          & 31        & 3.1          & 15        & 5.0          & 8          & 4.0           & 4.05                           \\ \hline
MiniGPT-4 (Vicuan 13B)          & 26       & 2.6          & 23        & 2.3          & 9         & 3.0          & 5          & 2.5           & 2.60                           \\
MiniGPT-4 (Vicuan 7B)          & 23       & 2.3          & 21        & 2.1          & 9         & 3.0          & 5          & 2.5           & 2.48                           \\
LLaVA (Vicuan 13B)          & 25       & 2.5          & 24        & 2.4          & 9         & 3.0          & 4          & 2.0           & 2.60                           \\ \hdashline[0.5pt/5pt]
ArtGPT-4  (Backbones MiniGPT-4-Vicuna-7B)         & 35       & 3.5          & 24        & 2.4          & 12        & 4.0          & 8          & 4.0           & 3.78                  \\ 
ArtGPT-4  (Backbones LLaVA-Vicuna-13B)         & 38       & 3.8          & 25        & 2.5          & 15        & 5.0          & 8          & 4.0           & 3.83                  \\ 
ArtGPT-4  (Backbones MiniGPT-4-Vicuna-13B)         & 38       & 3.8          & 28        & 2.8          & 15        & 5.0          & 8          & 4.0           & \textbf{3.90}                  \\ 
\hline
\bottomrule
\end{tabular}}
\caption{\label{Result3}Scoring of model outputs using the GPT-4 based on ArtMM for the scoring rules in the Supplementary Materials. We consider these four metrics, including Image Depiction Capability (IDC), Image Sentiment Analysis Capability (ISAC), Image Content Recognition Capability (ICRC), and Multi- round Dialogue Image Understanding Capability (MDIUC) to be equally important, and the total average is calculated as the mean of the average scores of the four metrics.}
\end{table*}

\subsection{Artistic-Understanding Evaluation}

\noindent\textbf{Evaluation on ArtEmis and ArtEmis-v2.0.}~~Evaluation results for artistic-understanding are presented in Table~\ref{Reasult}. We evaluate our proposed method against six state-of-the-art multimodal models on two art image explanation datasets: ArtEmis and ArtEmis-v2.0. About our proposed ArtGPT-4, all experiments utilized models pre-trained by either MiniGPT-4 or LLaVA, with training settings drawn from the Training section. From our results, we can observe that: 1) ArtGPT-4 consistently outperforms baseline models across all evaluation metrics on both ArtEmis and ArtEmis-v2.0 datasets. Specifically, the ArtGPT-4 model (Backbones on MiniGPT-4-Vicuna-13B) sets a new performance benchmark, outstripping the original MiniGPT-4-Vicuna-13B by a considerable margin. For instance, on the ArtEmis dataset, the ArtGPT-4 variant achieved VADER, TextBlob, and BERT scores of 0.813, 0.247, and 0.693 respectively. On the ArtEmis-v2.0 dataset, the scores were 0.987, 0.360, and 0.698 respectively, showcasing its superior performance. Furthermore, for different LLMs like MiniGPT-4-Vicuna-7B and Alpaca-7B, ArtGPT-4 attained VADER scores of 0.740 and 0.721, respectively, showcasing significant enhancements over the original models. 2) pitted against other state-of-the-art multimodal models, ArtGPT-4 (Backbones on LLaVA) surpassed the original LLaVA with 13B parameters, achieving VADER, TextBlob, and BERT scores of 0.799, 0.245, and 0.691 on the ArtEmis dataset. The model also outperformed the Multi-modal GPT, which had 1.66B parameters, despite only having 0.56B updated parameters. Most impressively, ArtGPT-4 (Backbones on MiniGPT-4-Vicuna-13B) eclipsed the performance of prior leading models such as VisualGPT, GIT, and ViLT. As a point of comparison, in the ArtEmis-v2.0 VADER metric, ArtGPT-4's score of 0.987 significantly outpaced VisualGPT's 0.141, GIT's 0.155, and ViLT's 0.082, reflecting a leap in performance by over sevenfold.

\noindent\textbf{Evaluation on ArtMM.} As illustrated in Table~\ref{Result3}, ArtGPT-4 (Backbones on MiniGPT-4-Vicuna-13B) boasts an impressive artistic-understanding ability with an average score of 3.90. This is markedly superior to the 2.60 average of the original MiniGPT-4 (Vicuna-7B) and is tantalizingly close to the 4.05 average achieved by human artists. For this comparison, we enlisted 10 artists (comprising 5 males and 5 females) to interpret images using the same guidelines provided to the model. In addition, other variants of ArtGPT-4 exhibited a notable average improvement when contrasted with other baseline models, including MiniGPT-4-Vicuna-7B and LLaVA-Vicuna-13B.

\begin{table*}[t]
\resizebox{.95\hsize}{!}{
\begin{tabular}{l|cc|ccc|ccc}
\toprule
\hline
\multirow{2}{*}{Methods}             & \multicolumn{1}{l}{Pretraing} & \multicolumn{1}{l|}{Learnable} & \multicolumn{3}{c|}{ArtEmis} & \multicolumn{3}{c}{ArtEmis-v2.0} \\
                                              & \multicolumn{1}{l}{Language model}                & parameters (\textbf{B}illion)            & VADER   & TextBlob  & BERT   & VADER    & TextBlob    & BERT    \\ \hline

\multicolumn{1}{l|}{MiniGPT-4+Only Fine-tune Mapping Modules} & Vicuna-13B                    & 0.003B                          & 0.746   & 0.242     & 0.692  & 0.939    & 0.332       & 0.674   \\
\multicolumn{1}{l|}{+Fine-tune LLM Top 5 layers} & Vicuna-13B                    & 1.58B                          & 0.750   & 0.242     & 0.681  & 0.944    & 0.333       & 0.673   \\
\multicolumn{1}{l|}{+Fine-tune all LLM (2 hours)}            & Vicuna-13B                    & 13B                            & 0.752   & 0.242     & 0.680  & 0.945    & 0.332       & 0.674   \\
\multicolumn{1}{l|}{+Fine-tune all LLM}            & Vicuna-13B                    & \underline{13B}                            & 0.815   & 0.250     & 0.693  & 0.988    & 0.359       & 0.695   \\
\hdashline[0.5pt/5pt]
ArtGPT-4 (Backbones MiniGPT-4)                                    & Vicuna-13B                    & \textbf{0.52B}                          & \textbf{0.813}   & \textbf{0.247}     & \textbf{0.693}  & \textbf{0.987}    & \textbf{0.360}       & \textbf{0.698}   \\
 \hline

\multicolumn{1}{l|}{MiniGPT-4+Only Fine-tune Mapping Modules}            & Vicuna-7B                    & 0.003B                            & 0.705          & 0.225          & 0.683          & 0.911          & 0.322          & 0.660   \\
\multicolumn{1}{l|}{+Fine-tune LLM Top 5 layers}            & Vicuna-7B                    & 1.01B                            & 0.710          & 0.227          & 0.686          & 0.914          & 0.322          & 0.664   \\ 
\multicolumn{1}{l|}{+Fine-tune all LLM}            & Vicuna-7B                    & \underline{7B}                            & 0.740          & 0.234          & 0.686          & 0.921          & 0.326          & 0.665   \\
\hdashline[0.5pt/5pt]
ArtGPT-4 (Backbones MiniGPT-4)                                    & Vicuna-7B  &                  \textbf{0.26B}                          & \textbf{0.740} & \textbf{0.233} & \textbf{0.686} & \textbf{0.920} & \textbf{0.327} & \textbf{0.665}   \\
 \hline

\multicolumn{1}{l|}{MiniGPT-4+Only Fine-tune Mapping Modules}            & Alpaca-7B                    & 0.003B                            & 0.681          & 0.221          & 0.670          & 0.853          & 0.318          & 0.649   \\ 
\multicolumn{1}{l|}{+Fine-tune LLM Top 5 layers}            & Alpaca-7B                    & 1.01B                            & 0.683          & 0.222          & 0.670          & 0.855          & 0.320          & 0.650   \\
\multicolumn{1}{l|}{+Fine-tune all LLM}            & Alpaca-7B                    & \underline{7B}                            & 0.722          & 0.230          & 0.680          & 0.920          & 0.328          & 0.650   \\
\hdashline[0.5pt/5pt]
ArtGPT-4 (Backbones MiniGPT-4)                                    & Alpaca-7B  &                  \textbf{0.26B}                          & \textbf{0.721} & \textbf{0.233} & \textbf{0.685} & \textbf{0.923} & \textbf{0.326} & \textbf{0.669}   \\
\hline

LLaVA ~\citep{liu2023visual}                                               & Vicuna-13B                    & \underline{13B}                            & 0.740   & 0.240     & 0.688  & 0.935    & 0.332       & 0.673   \\

+Fine-tune all LLM in Stage 1                                              & Vicuna-13B                    & \underline{13B}                            & 0.800   & 0.245     & 0.690  & 0.987    & 0.352       & 0.692   \\ \hdashline[0.5pt/5pt]
ArtGPT-4 (Backbones LLaVA)                                    & Vicuna-13B  &                  \textbf{0.52B}                          & \textbf{0.799} & \textbf{0.245} & \textbf{0.691} & \textbf{0.982} & \textbf{0.350} & \textbf{0.689}   \\
\hline
\bottomrule
\end{tabular}}
\caption{\label{Reasult_e}Effectiveness of proposed components. We compare to baselines on ArtEmis and ArtEmis-v2.0 datasets.}
\end{table*}

\begin{table*}[t]
\centering
\resizebox{.95\hsize}{!}{
\begin{tabular}{l|c|cccccc}
\toprule
\hline
\multirow{2}{*}{Methods} & \multicolumn{1}{l|}{Trainable} & \multicolumn{3}{c|}{ArtEmis}                  & \multicolumn{3}{c}{ArtEmis-v2.0} \\
                                  & parameters (Billion)           & VADER & TextBlob & \multicolumn{1}{c|}{BERT}  & VADER    & TextBlob    & BERT    \\ \hline
ArtGPT-4 (Backbones MiniGPT-4-Vicuna-13B)                         & 0.52B                          & 0.813 & 0.247    & \multicolumn{1}{c|}{0.693} & 0.987    & 0.360       & 0.698   \\
- Freeze all RMS Norm (before I-MHA)     & 0.52B                          & 0.801 & 0.239    & \multicolumn{1}{c|}{0.690} & 0.977    & 0.358       & 0.696   \\
+ Train  all RMS Norm (before I-MHA)     & 0.52B                          & 0.813 & 0.247    & \multicolumn{1}{c|}{0.693} & 0.987    & 0.360       & 0.698   \\
- Freeze all RMS Norm (following Image Adapter)        & 0.52B                          & \multicolumn{6}{c}{- (Vanishing gradient)}                                      \\
- Remove all Image Adapter                   & 0.0004B                        & 0.746 & 0.242    & \multicolumn{1}{c|}{0.693} & 0.939    & 0.333       & 0.673   \\
- Remove 1/2 Image Adapter               & 0.26B                          & 0.771 & 0.243    & \multicolumn{1}{c|}{0.689} & 0.945    & 0.338       & 0.688   \\
- Remove 1/4 Image Adapter               & 0.14B                          & 0.747 & 0.240    & \multicolumn{1}{c|}{0.688} & 0.940    & 0.333       & 0.671   \\
+ Add 1/2 Image Adapter               & 0.78B                          & 0.820 & 0.247    & \multicolumn{1}{c|}{0.694} & 0.988    & 0.360       & 0.698   \\
+ Add 1/4 Image Adapter               & 0.65B                          & 0.813 & 0.247    & \multicolumn{1}{c|}{0.693} & 0.987    & 0.360       & 0.698   \\ \hline
\bottomrule
\end{tabular}}
\caption{\label{ablation}Scores of ablation experiments for each module on the dataset.}
\end{table*}

\subsection{Evaluations on Components}

\noindent\textbf{Backbone via MiniGPT-4-Vicuna-13B.}~~We conducted four experiments using MiniGPT-4-Vicuna-13B as backbones. 1) While only the model mapping module was fine-tuned, we observed scores of 0.746, 0.242, and 0.692 on VADER, TextBlob, and BERT metrics respectively for ArtEmis. 2) By fine-tuning the initial five layers of the LLM, the scores slightly increased to 0.750, 0.242, and 0.681. 3) When the entire LLM was fine-tuned over a span of two hours, the scores were 0.752, 0.242, and 0.680. 4) While the entire LLM was fine-tuned without any time constraints, we saw significant improvements with scores reaching 0.815, 0.250, and 0.693. ArtGPT-4 (Backbones on MiniGPT-4-Vicuna) exhibited the most robust performance in all groups, especially in the fourth. Notably, with just two hours of fine-tuning on 0.56B parameters, it demonstrated performance comparable to unrestricted fine-tuning of the LLM's 13B parameters.

\noindent\textbf{Backbone via MiniGPT-4-Vicuna-7B and -Alpaca-7B.}~~For MiniGPT-4-Vicuna-7B, the scores were 0.705, 0.225, and 0.683 in the first experiment, which improved to 0.710, 0.227, and 0.686 in the second. In the third experiment, the scores reached 0.740, 0.234, and 0.686. On the other hand, with the MiniGPT-4 Alpaca-7B model, we observed scores of 0.681, 0.221, and 0.670 in the first experiment. These scores slightly increased to 0.683, 0.222, and 0.670 in the second, and further to 0.722, 0.230, and 0.680 in the third experiment. Impressively, our ArtGPT-4, trained on only 0.26B parameters, achieved significant performance gains in both MiniGPT-4-Vicuna-7B and Alpaca-7B models, especially on the ArtEmis dataset. The ArtGPT-4 model even surpassed the results of the full fine-tuning MiniGPT-4-Alpaca model.

\noindent\textbf{Backbone via LLaVA-Vicuna-13B.}~~We investigated the performance of models backbones on the LLaVA architecture pre-trained with the Vicuna-13B language model. Initially, the LLaVA model without any further fine-tuning achieved VADER, TextBlob, and BERT scores of 0.740, 0.240, and 0.688 respectively on the ArtEmis dataset. In a subsequent experiment, the entire LLM underwent fine-tuning in stage 1. This resulted in improved scores, where VADER reached 0.800, TextBlob was at 0.245, and BERT scored 0.690. This demonstrated the potential enhancements achievable with fine-tuning. When we observed the ArtGPT-4 model, backbones on the LLaVA architecture, and updated with only an additional 0.52B parameters, it showcased VADER, TextBlob, and BERT scores of 0.799, 0.245, and 0.691 respectively. These results not only surpassed the original LLaVA model's performance by notable margins across all metrics but also closely matched the performance of the extensively fine-tuned LLMs in the second group, which employed a whopping 13B parameters. This illustrates the efficiency and potential of the ArtGPT-4 model in leveraging smaller parameter updates for significant performance gains.

\subsection{Ablation}

We present the results of our ablation experiments conducted to analyze the impact of different modules on the performance of ArtGPT-4. Table \ref{ablation} displays the scores obtained for each module on the ArtEmis and ArtEmis-v2.0 datasets. To investigate the effects of various components, we conducted ablations by modifying the model as follows: When we removed RMS normalization before the Image Multi-Head Attention (I-MHA) layer, the performance on VADER dropped from 0.813 to 0.801, while TextBlob's score slightly decreased from 0.247 to 0.239. However, BERT's score remained almost the same at 0.693. Turning the RMS Norm layer before I-MHA on for training produced identical results, with VADER at 0.813, TextBlob at 0.247, and BERT at 0.693. However, when RMS normalization was removed following the Image Adapter layer, the gradients vanished during training, leading to a lack of meaningful results. The Image Adapter plays a crucial role in the overall performance of ArtGPT-4. Completely removing the Image Adapter resulted in a drop in VADER to 0.746, TextBlob to 0.242, and BERT remained consistent at 0.693. When we removed half of the Image Adapter, the performance on VADER was 0.771, TextBlob was 0.243, and BERT was 0.689. Removing a quarter of the Image Adapter caused VADER to drop to 0.747, TextBlob to 0.240, and BERT to 0.688. On the other hand, adding half of the Image Adapter boosted the VADER score to 0.820, and adding a quarter resulted in a VADER score of 0.813. However, in both cases, TextBlob and BERT scores remained consistent at 0.247 and 0.693 respectively. Adding few Image Adapter parameters results in inadequate performance, while adding many does not yield significant improvements and can lead to an more parameters.

\section{Conclusion}
Our experimental results demonstrates significant progress of ArtGPT-4 in the field of vision-language understanding, showing superior performance to its predecessor, MiniGPT-4 or LLaVA. Our proposed modifications, including added adapter image layers, have optimized the model's performance and addressed the artistic-understanding challenges posed by vision-language tasks. Additionally, we have introduced a novel benchmark for evaluating the performance of vision-language models, which provides a more comprehensive criterion for assessing these models. Our model was trained in just 2 hours, using a relatively small dataset, and achieved state-of-the-art performance. And our training method can be applied to different multimodal models. This work effectively bridges the gap between art and LLM.

\bibliography{colm2024_conference}
\bibliographystyle{colm2024_conference}

\appendix
\newpage
\section{Stage 2 training instructions}

According to Table~\ref{ins}, we used the same instructions as MiniGPT-4 and LLaVA in order to ensure the fairness of our comparison.
\begin{table}[htbp]
\centering
\resizebox{\linewidth}{!}{
\begin{tabular}{l}
\hline
\textbf{LLaVA Instructions}                                   \\ \hline
Describe the following image in detail                        \\
Provide a detailed description of the given image             \\
Give an elaborate explanation of the image you see            \\
Share a comprehensive rundown of the presented image          \\
Offer a thorough analysis of the image                        \\
Explain the various aspects of the image before you           \\
Clarify the contents of the displayed image with great detail \\
Characterize the image using a well-detailed description      \\
Break down the elements of the image in a detailed manner     \\
Walk through the important details of the image               \\
Portray the image with a rich, descriptive narrative          \\
Narrate the contents of the image with precision              \\
Analyze the image in a comprehensive and detailed manner      \\
Illustrate the image through a descriptive explanation        \\
Examine the image closely and share its details               \\
Write an exhaustive depiction of the given image              \\ \hline
\textbf{MiniGPT-4 Instructions}                               \\ \hline
Describe this image in detail                                 \\
Take a look at this image and describe what you notice        \\
Please provide a detailed description of the picture          \\
Could you describe the contents of this image for me          \\ \hline
\end{tabular}}
\caption{It describes all the instructions trained in the second stage of LLaVA and MiniGPT-4. the second training stage of ArtGPT-4 is a random selection of these instructions.}~\label{ins}
\end{table}

\section{Zero-shot Testing Datasets}\label{Datasets}

\begin{figure*}[t]
    \centering
  \includegraphics[width=.9\textwidth]{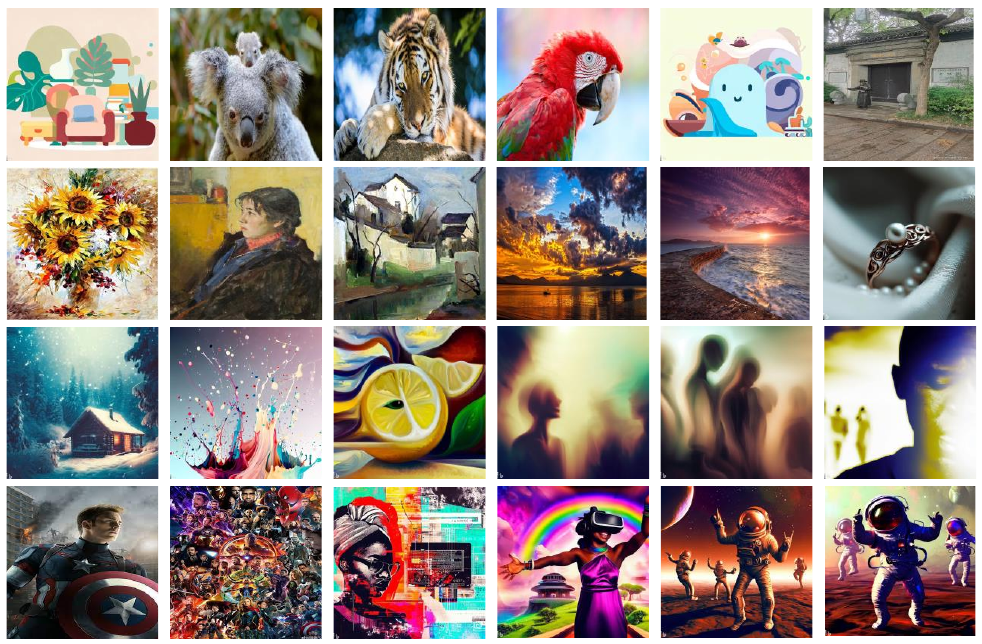}
  \caption{This image is a visual example of the ArtMM dataset, which is a compilation of various artworks and photographs. It includes a wide range of subjects like wildlife, abstract and traditional art, landscapes, and imaginative fantasy scenes. The collection also incorporates elements from popular culture and cinematic stills.}
  \vspace{-5pt}
  \label{fig:10}
\end{figure*}

\textbf{ArtEmis}~\citep{achlioptas2021artemis} delves into the intricate relationship between visual content, emotional impact, and language-based explanations. The data of This dataset with annotators indicating dominant emotions and providing grounded verbal explanations. ArtEmis comprises 455K emotion attributions and explanations on 80K artworks from WikiArt~\citep{artgan2018}. \textbf{ArtEmis-v2.0}~\citep{youssef2022artemis2} builds upon the original ArtEmis dataset by employing a novel contrastive data collection approach. By balancing emotional biases and incorporating 260,533 new instances with contrasting emotions, the dataset achieves a more fine-grained representation of emotions and associated painting explanations. Furthermore, to further enrich our evaluation, we filtered images from the mPLUG-Owl~\citep{ye2023mplugowl} database and others, in total the 40 image-instruction data called \textbf{ArtMM}, the example as shown in Figure~\ref{fig:10}. These images, characterized by their complex elements, were a mix of those found online and others generated using DALL-E 2~\citep{ramesh2022hierarchical}. These images were chosen through a meticulous process involving 10 artists (comprising an equal gender distribution of 5 males and 5 females), who selected the images based on artistic scoring (realism) and the ability to evoke emotional expression.

\section{Evaluation Metrics}\label{Metrics}

\noindent\textbf{VADER.}~~ For VADER-based similarity~\citep{hutto2014vader}, we utilized the VADER sentiment analyzer to compute the compound sentiment score for each response. The compound sentiment score represents the overall sentiment of the response, ranging from -1 (most negative) to 1 (most positive). The sentiment similarity between two responses, model response and Labeling of data sets, was then calculated as the absolute difference between their respective compound sentiment scores. A higher score indicates better performance.

\noindent\textbf{TextBlob.}~~ For TextBlob-based similarity~\citep{loria2018textblob},, we used the TextBlob library to analyze the polarity (sentiment score) of each response. The polarity ranges from -1 (most negative) to 1 (most positive). Similar to VADER-based similarity, the sentiment was computed as the absolute difference between their polarity scores. Higher scores indicate better performance.

\noindent\textbf{BERT.}~~ For BERT-based similarity, we employed the SentenceTransformer BERT~\citep{reimers2019sentence} model to encode text into high-dimensional embeddings. We then utilized the cosine similarity metric to quantify the similarity between their embeddings. The cosine similarity, as shown in 
Equation (\ref{COS}) ranges from -1 (completely dissimilar) to 1 (identical). Higher scores indicate better performance.

\begin{equation} \label{COS}
    \cos \theta =\frac{  \mathbf{Y}_{A} \cdot \mathbf{Y}_{B}}{\left \| \mathbf{Y}_{A} \right \| \times \left \| \mathbf{Y}_{B} \right \|}
\end{equation}
Where $\mathbf{A}$ denotes the word embedding vector of the model responses and $\mathbf{B}$ denotes the word embedding vector of the labels in the dataset. The $\mathbf{A} \cdot \mathbf{B}$ denotes the dot product of vectors $\mathbf{A}$ and $\mathbf{B}$ and $\left \| \cdot \right \|$ denotes the Euclidean norm.

\section{Quality Comparison}


\paragraph{Image Description:} Both MiniGPT-4 and ArtGPT-4 provide detailed descriptions of the images they are given as shown in Table 2, describing a traditional Chinese ink painting of a landscape scene. First, ArtGPT-4 provides more detailed and specific information about the subject matter of the painting, including details about the rocks, trees, and grasses depicted in the image. Second, ArtGPT-4 makes use of more specialized language to describe the painting, including terms such as "Chinese calligraphy brushstrokes" and "shading and texture of the rocks." This indicates a greater level of expertise and familiarity with the subject matter, which can help to lend credibility and authority to the description. Finally, ArtGPT-4 uses language that emphasizes the emotional and aesthetic impact of the painting, describing it as having a "tranquil" and "beautiful" effect that emphasizes the rugged, wild beauty of the natural landscape. This suggests that ArtGPT-4 is better able to understand and convey the emotional and aesthetic impact of visual art, which could be particularly useful in fields such as art criticism or curation.

\paragraph{Aesthetics:}Both MiniGPT-4 and ArtGPT-4 recognized the artistic qualities of the image as shown as shown in Table 3, but ArtGPT-4's response is more detailed and descriptive. ArtGPT-4 not only describes the image but also provides an interpretation of it, highlighting the beauty in decay and evoking emotions such as sadness, loneliness, and desolation. In addition, ArtGPT-4 provides a more technical analysis of the image, discussing the composition, lighting, color palette, and tonal range. These details demonstrate a deeper understanding of the elements of visual arts and photography. Overall, ArtGPT-4's response is more nuanced and insightful, showcasing its superior capacity for understanding and analyzing art.

\paragraph{Better-Looking Websites:}ArtGPT-4's response includes an image that serves as a visual representation of the joke website as shown in Table 4. The use of the image adds an extra layer of creativity to the website and can help to attract visitors. Additionally, the CSS styling used in ArtGPT-4's response is more comprehensive, providing more visual appeal to the website. In comparison, MiniGPT-4's response uses a more basic CSS styling and does not include an image. The website still looks functional and readable, but it lacks the same level of visual interest that ArtGPT-4's response provides. Overall, ArtGPT-4's response demonstrates a more sophisticated understanding of web design and has more visual appeal than MiniGPT-4's response.

\section{ArtMM Benchmarks}\label{Bench}
Similar to how the TOEFL and IELTS tests are used to measure English language proficiency, we were enthusiastic about establishing a reliable standard for evaluating the ability to understand multimodal images. We implemented four scoring criteria and a five-point scoring scale to evaluate the model's capacity for understanding images.

\paragraph{Image Depiction Capability (IDC):}We selected 10 various types of graphs, such as paintings, photographs, AI-generated images, etc., for the model to provide descriptions for. Each image is scored according to the following criteria:

\begin{itemize}
\item 0: No image description capability
\item 1: Description does not match real image representation
\item 2: Partial image description
\item 3: Complete image description without appreciation information
\item 4: Complete image description at the human level of appreciation
\item 5: Complete image description surpassing the human level of appreciation, such as an artist.
\end{itemize}

\paragraph{Image Sentiment Analysis Capability (ISAC):} We chose 10 images of individuals and instructed the participants to "Analyze the emotions expressed by the individuals in the images as well as the emotions felt by the viewer observing them." Each image is scored according to the following criteria:

\begin{itemize}
\item 0: Can't describe the feelings about the picture
\item 1: Can describe the relevant emotion but no logical proof (e.g.: the picture is seen... So people will have a kind of... emotion)
\item 2: Can describe the relevant emotion and justify it. But the description is not perfect
\item 3: The individuals in the images or the viewer's emotions can be described perfectly and justified.
\item 4: Can describe all the emotions as an ordinary person and justify them.
\item 5: Can describe all emotions perfectly and justifiably and full of art.
\end{itemize}

\paragraph{Image Content Recognition Capability (ICRC):} We selected 5 images with a variety of objects and scenes, such as animals, landscapes, and household items, for the model to recognize and label. Each image is scored according to the following criteria:

\begin{itemize}
\item 0: No image content recognition capability
\item 1: Some objects/scenes are recognized but with significant errors or omissions
\item 2: Most objects/scenes are recognized with some errors or omissions
\item 3: All objects/scenes are recognized with few errors or omissions
\item 4: All objects/scenes are recognized with high accuracy and speed, comparable to a human observer
\item 5: All objects/scenes are recognized with high accuracy, speed, and contextual understanding, surpassing the performance of a human observer.
\end{itemize}

\paragraph{Multi-round Dialogue Image Understanding Capability (MDIUC):} We randomly select 2 images and conduct five rounds of dialogue with the model for each image to assess its multi-round image understanding capability. Each image-dialogue pair is scored according to the following criteria:

\begin{itemize}
\item 0: No image understanding capability in the dialogue
\item 1: Partial image understanding, but unable to carry on the dialogue smoothly
\item 2: Able to understand the image to some extent and carry on the dialogue with some coherence, but lacks understanding of some key points
\item 3: Can understand the image and carry on the dialogue smoothly, but with some minor misunderstandings or mistakes
\item 4: Can understand the image and carry on the dialogue smoothly, with accurate understanding and good coherence
\item 5: Can understand the image and carry on the dialogue smoothly, with accurate understanding, good coherence, and creative responses.
\end{itemize}

We incorporated these scoring criteria into the state-of-the-art GPT-4 model to establish a comprehensive multimodal model evaluation system.

\begin{table*}[htbp]
\centering
\begin{tabular}{|lp{0.8\textwidth}|}
\hline
\multicolumn{2}{|l|}{\textbf{ArtGPT-4 and image example compare}} \\
\hline
\textbf{Ures} & Could you please describe the diagram? \\
 & \includegraphics[width=5cm]{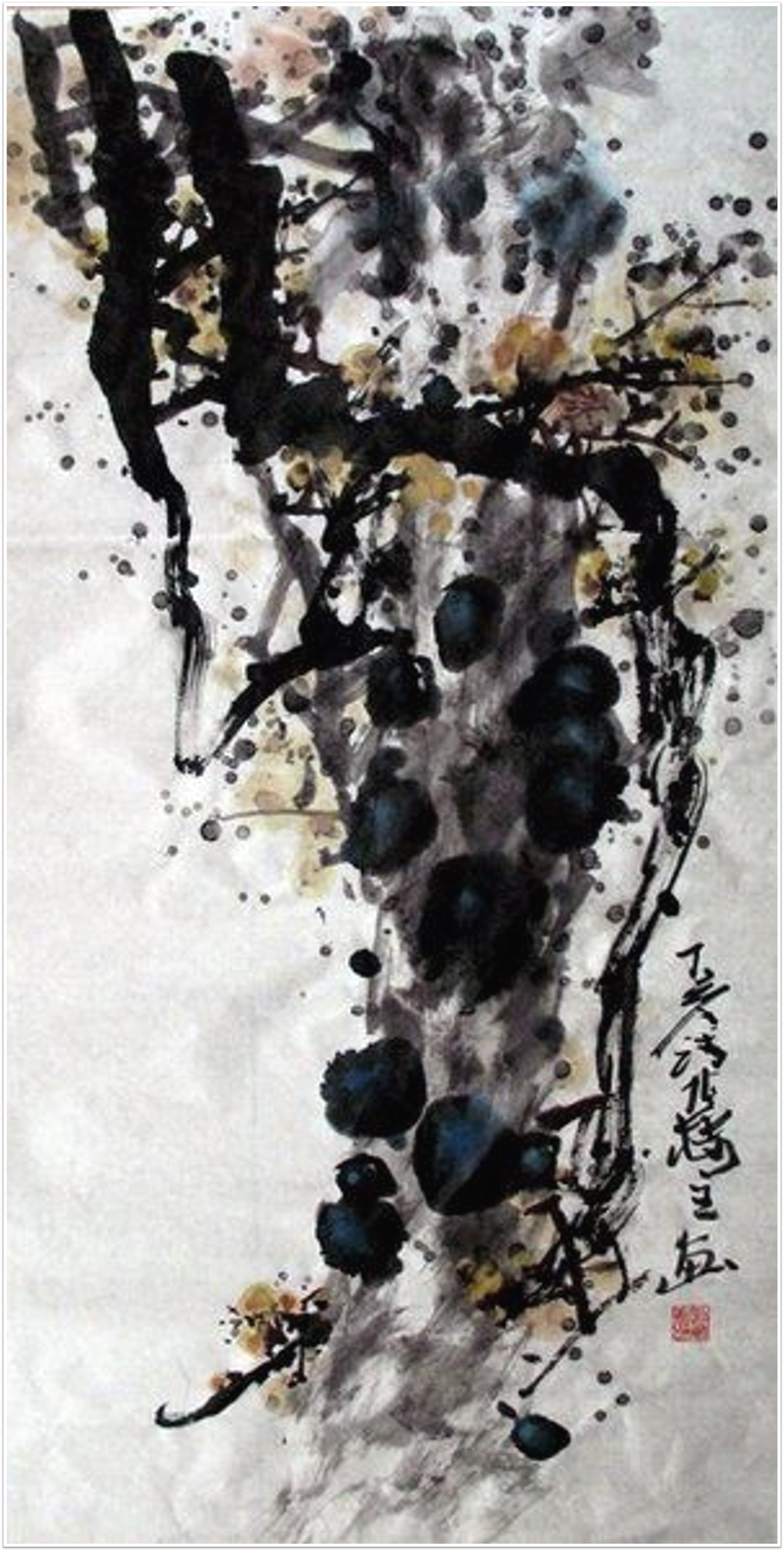} \\
\hline
\textbf{MiniGPT-4} & The image appears to be an abstract ink painting with black and white keys brush strokes creating various shapes and patterns on a white background. 
Some of the shapes appear to be tree branches, while others are more geometric in shape. 
There are also some splashes of white ink that appear to be floating in the air. 
The overall effect is one of movement and energy, with the brush strokes creating a sense of motion and flow.
 \\
 
\hline
\textbf{ArtGPT-4} & The image depicts a traditional Chinese ink painting. The painting features a group of rocks and boulders on a hill, with black ink used to create the shading and details of the rocks. The overall effect is one of rugged, natural beauty, with the black ink contrasting against the white paper to create a striking visual impact. 

The rocks are depicted in various sizes, with some larger and more prominent than others, and the overall composition creates a sense of depth and space. 
The use of Chinese calligraphy brushstrokes is evident in the way the ink is applied, with thin, delicate lines used to depict the trees and grasses, and thicker, bolder strokes used to create the shading and texture of the rocks.
The overall effect is one of tranquility and beauty, with the natural landscape depicted in a way that emphasizes its rugged, wild beauty. 

The use of black ink creates a sense of contrast and depth, with the dark tones of the ink contrasting against the white paper to create a striking visual impact. The rocks are depicted in various sizes, with some larger and more prominent than others, and the overall composition creates a sense of depth and space. 

The use of Chinese calligraphy brushstrokes is evident in the way the ink is applied, with thin, delicate lines used to depict the trees and grasses, and thicker, bolder strokes used to create the shading and texture of the rocks. 
The overall effect is one of tranquility and beauty, with the natural landscape depicted in a way that emphasizes its rugged, wild beauty.
\\
\hline
\end{tabular}
\caption{The image description capability of GPT-4 compared with the ArtGPT-4. ArtGPT-4 provides more detailed and specific information about the subject matter of the painting. ArtGPT-4 uses language that emphasizes the emotional and aesthetic impact of the painting, describing it as having a "tranquil" effect that emphasizes the rugged, wild beauty of the natural landscape.}
\label{Art1}
\end{table*}

\begin{table*}[htbp]
\centering
\begin{tabular}{|lp{0.8\textwidth}|}
\hline
\multicolumn{2}{|l|}{\textbf{ArtGPT-4 and image example compare}} \\
\hline
\textbf{Ures} & Do you think this picture is artistic? Why? \\
 & \includegraphics[width=6cm]{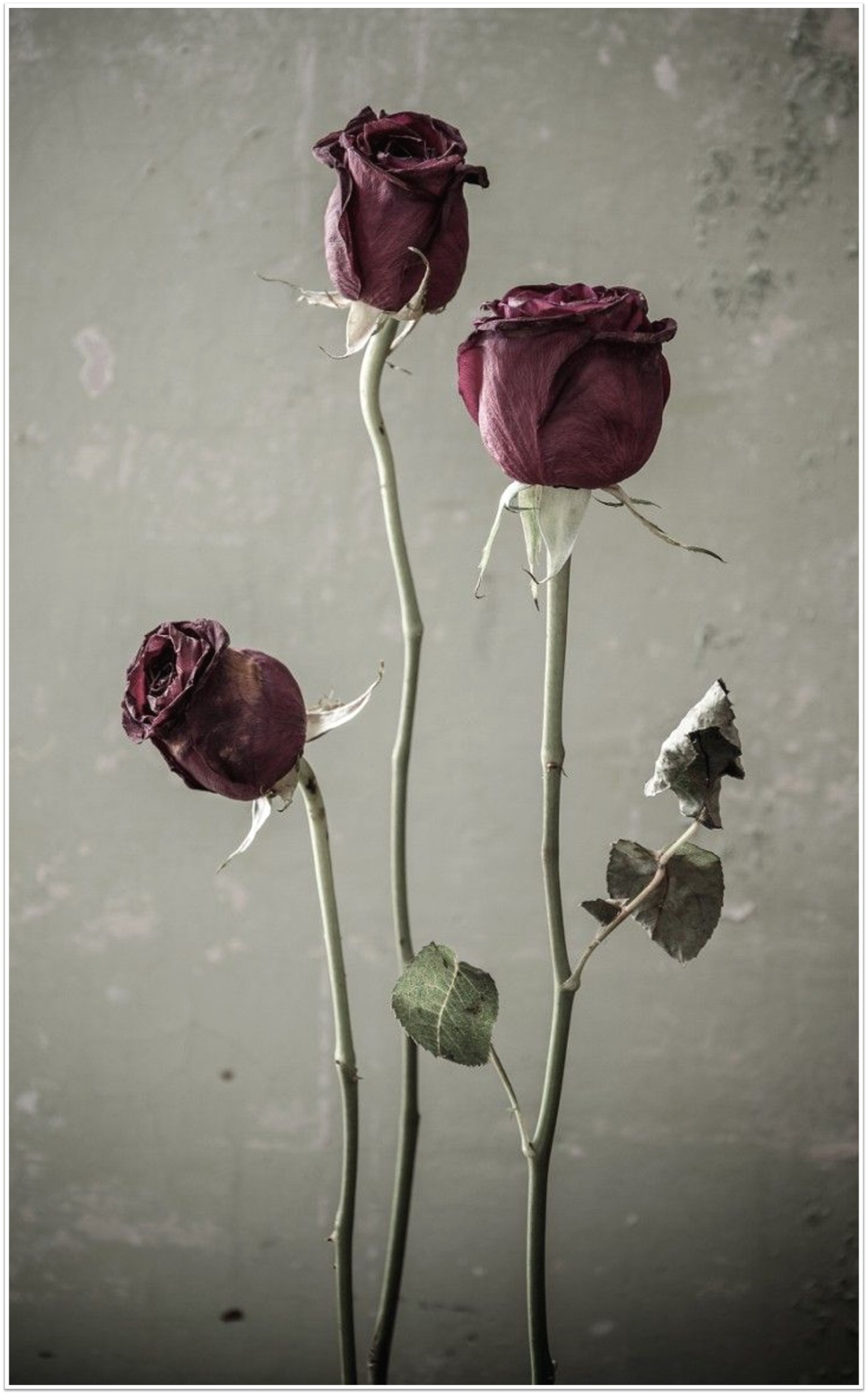} \\
\hline
\textbf{MiniGPT-4} & Yes, I think this image is artistic. The image appears to be a close-up shot of three roses, with one in the foreground and the other two in the background. The lighting is dim, with shadows cast on the petals and stems of the roses. The colors are muted, with the roses appearing to be a deep, dark red. The overall effect is one of stillness and quiet, with the roses appearing to be suspended in time.
 \\
\hline
\textbf{ArtGPT-4} & Yes, I think this image is artistic. The photograph captures the essence of beauty in decay, the fading colors of the petals, and the empty space around the roses. The image evokes feelings of sadness, loneliness, and desolation.

The composition of the photograph is well balanced, with the roses in the foreground and the background fading into greyscale. The lighting creates a dramatic effect, highlighting the decay of the flowers and casting deep shadows on the walls.

The color palette is rich and warm, with the red roses standing out against the fading grey wall. The tonal range is well-balanced, with the roses’ deep red contrasting with the lighter tones of the petals and the wall.

Overall, the image is well-executed and captures the beauty in decay. It is an example of still life photography, capturing the fleeting beauty of nature and preserving it in a way that emphasizes its transience.

\\
\hline
\end{tabular}
\caption{The Aesthetics description capability of GPT-4 compared with the ArtGPT-4. ArtGPT-4 not only describes the image but also provides an interpretation of it, highlighting the beauty in decay and evoking emotions such as sadness, loneliness, and desolation. In addition, ArtGPT-4 provides a more technical analysis of the image.}
\label{Art1}
\end{table*}

\begin{table*}[htbp]
\centering
\begin{tabular}{|lp{0.8\textwidth}|}
\hline
\multicolumn{2}{|l|}{\textbf{ArtGPT-4 and image example compare}} \\
\hline
\textbf{Ures} & Can you please give me an executable HTML and CSS code and spruce up this site?\\
 & \includegraphics[width=10cm]{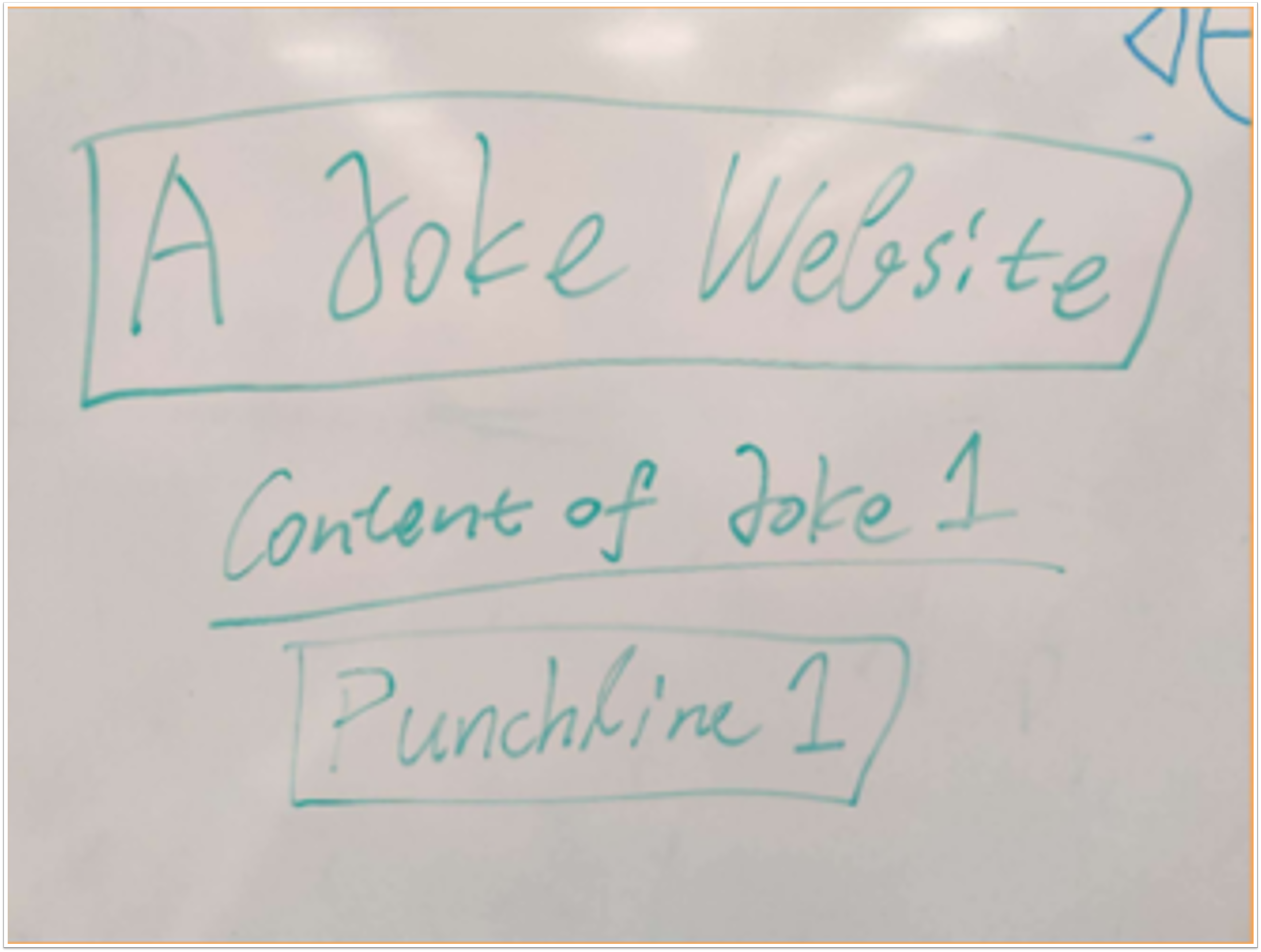} \\
\hline
\textbf{MiniGPT-4} &
\includegraphics[width=10cm]{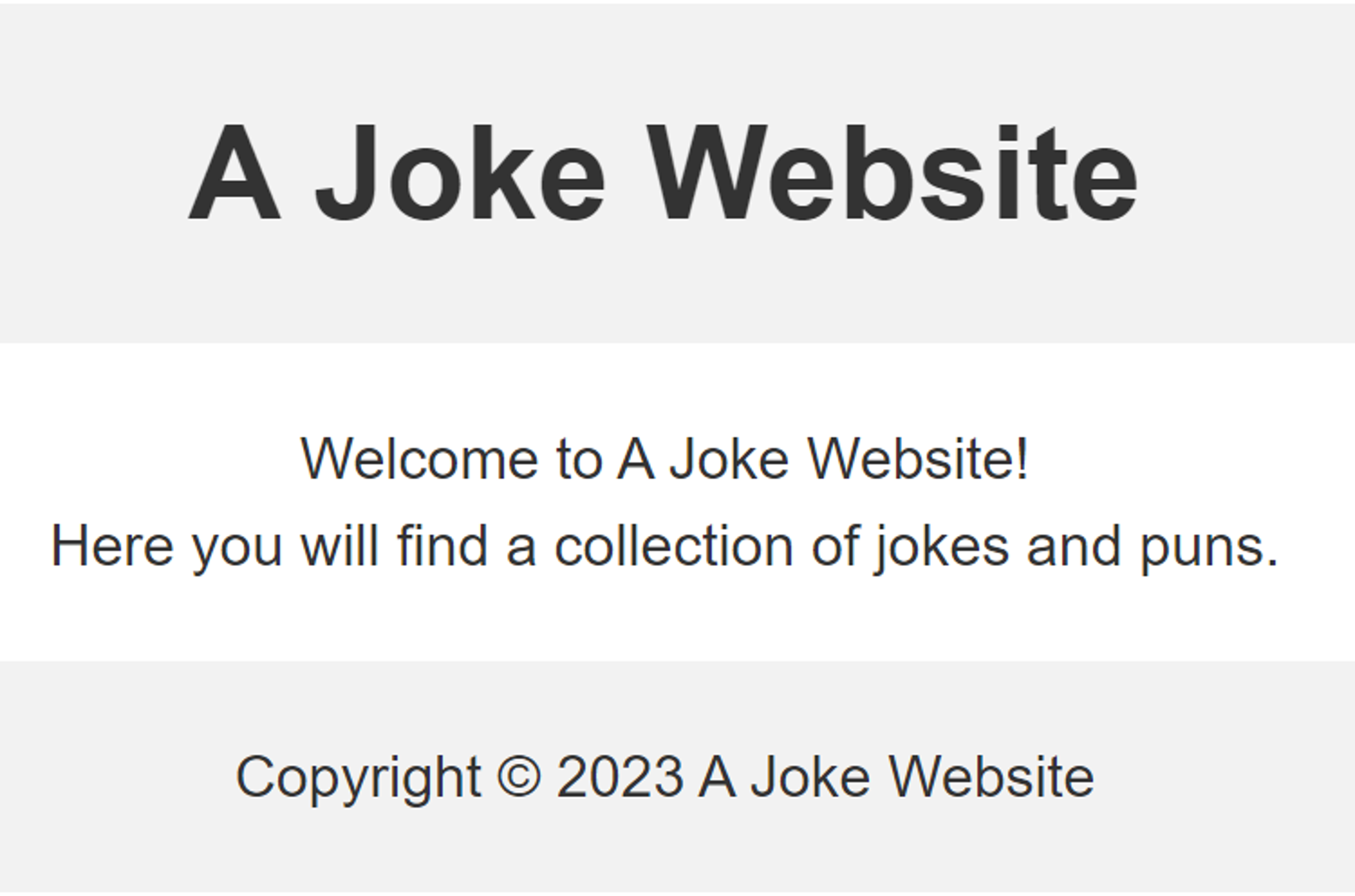}

 \\
\hline
\textbf{ArtGPT-4} & 
\includegraphics[width=10cm]{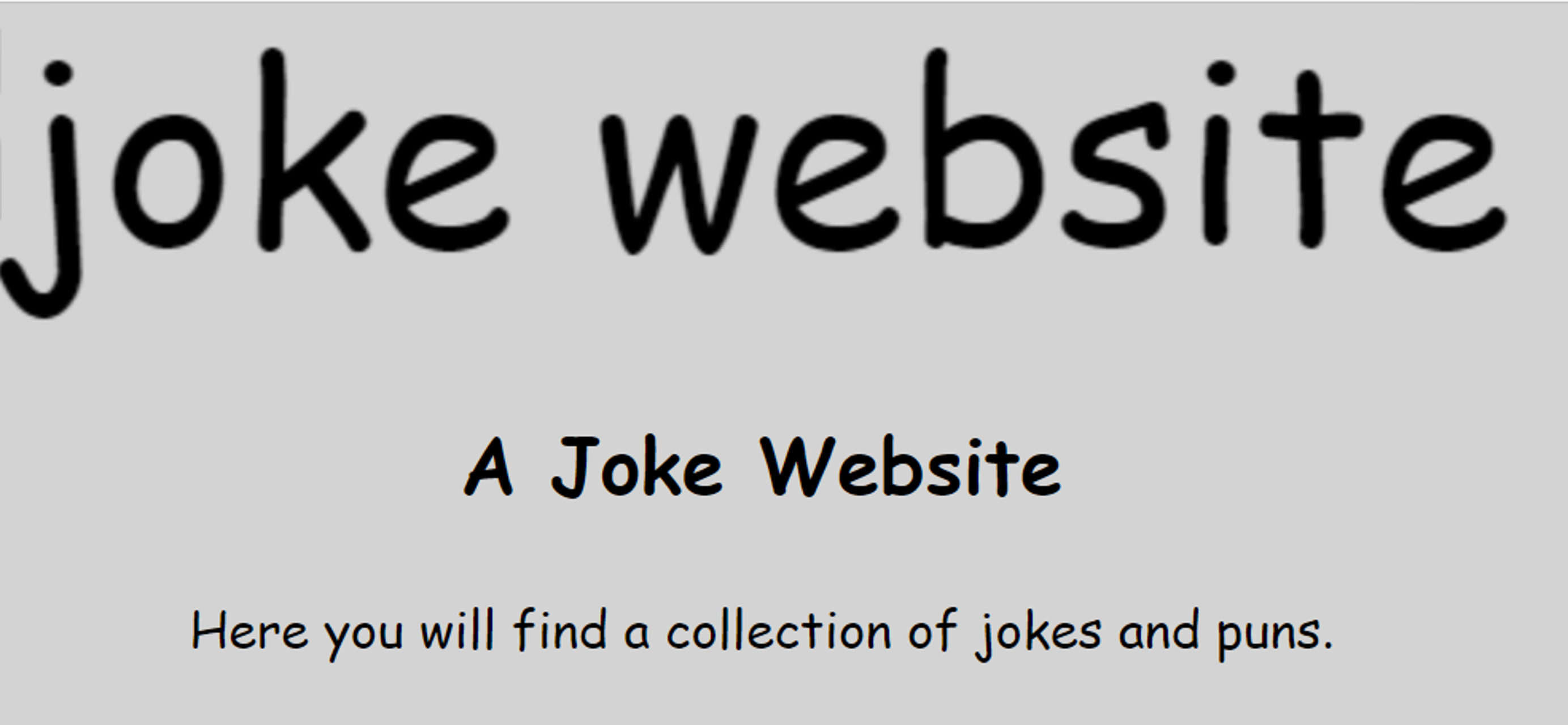}
\\
\hline
\end{tabular}
\caption{ArtGPT-4 use of the image adds an extra layer of creativity to the website and can help to attract visitors. Additionally, the CSS styling used in ArtGPT4’s response is more comprehensive, providing more visual appeal to the website. In comparison, MiniGPT-4’s response uses a more basic CSS styling and does not include an image. The website still looks functional and readable, but it lacks the same level of visual interest that ArtGPT-4’s response provides. }
\label{Art1}
\end{table*}

\section{Compare with GPT-4}
The following four examples demonstrate that ArtGPT-4 possesses comparable or even superior image comprehension abilities compared to GPT-4. This achievement is a remarkable breakthrough, given that ArtGPT-4 has only 13 billion parameters, significantly fewer than GPT-4's 175+ billion parameters.

\subsection{Image Depiction Capability}
\begin{table*}[htbp]
\centering
\begin{tabular}{|lp{0.8\textwidth}|}
\hline
\multicolumn{2}{|l|}{\textbf{ArtGPT-4 and image example compare}} \\
\hline
\textbf{Ures} & Here is an audience submitted image \\
 & \includegraphics[width=6cm]{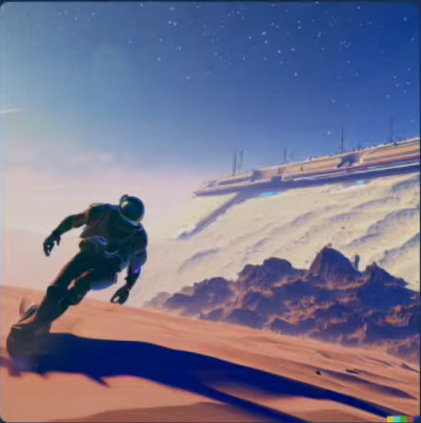} \\
\hline
\textbf{GPT-4} & This image depicts an astronaut exploring a snowy, mountainous landscape on an alien planet. In the background, we see a large spacecraft resting on the snow-covered ground, with its antennas and various equipment. The sky above is filled with stars, and a celestial body can be seen in the distance. The astronaut appears to be wearing a green spacesuit and is captured in motion, perhaps making his way towards the spacecraft. \\
\hline
\textbf{ArtGPT-4} & This is a screen shot of a video game or virtual environment. The image shows a desert landscape with mountains in the background, and a person wearing a spacesuit standing on the ground with their arms outstretched. It looks like they are running or jumping, but it is not clear from this image alone. The image has a clear and detailed representation of the environment, with realistic textures and lighting. The colors are vibrant and contrasting, making the image look very appealing. The overall quality of the image is high, with excellent details and contrast.\\
\hline
\end{tabular}
\caption{In the example video (\url{https://www.youtube.com/watch?v=outcGtbnMuQ}) the image description capability of GPT-4 compared with the ArtGPT-4}
\label{Art1}
\end{table*}

As shown in Table~\ref{Art1}, ArtGPT-4's description of the image is highly effective in conveying a sense of action and excitement. The desert landscape and the figure in the space suit are rendered in vibrant detail, with expert use of texture, lighting and colour. It is successful in creating a tangible sense of atmosphere and energy. The description's emphasis on the image's artistic merits, such as its vibrant colours and realistic textures, provides a vivid impression of the image as a digital artwork.

\subsection{Understanding Human Emotional Capabilities}

\begin{table*}[htbp]
\centering
\begin{tabular}{|lp{0.8\textwidth}|}
\hline
\multicolumn{2}{|l|}{\textbf{ArtGPT-4 and image example compare}} \\
\hline
\textbf{Ures} & What’s funny about this image? \\
 & \includegraphics[width=6cm]{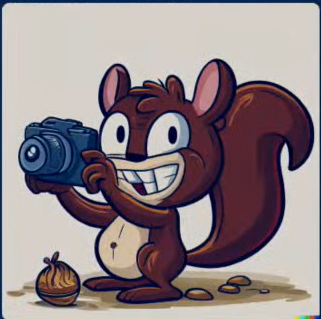} \\
\hline
\textbf{GPT-4} & The image is funny because it shows a squirrel holding a camera and taking a photo of a nut as if it were a professional photographer. let's humorous situation because squirrels typically eat nuts, and we don't expect them to use a camera or act like humans, \\
\hline
\textbf{ArtGPT-4} & This image is funny because it depicts a cartoon squirrel holding a camera with a nut in its hand. The squirrel is wearing a camera strap and looking at the camera with a smile on its face. It is a cute and playful image that can make people laugh.

The squirrel’s facial expression is funny because it is playful and expressive. The fact that the squirrel is holding a camera while eating nuts is also amusing because it shows that the animal is multitasking and is not afraid to try new things.

Overall, this image is funny because it is a cute and playful representation of a squirrel doing something unexpected, which is taking a photo with a camera while eating nuts.\\
\hline
\end{tabular}
\caption{In the example video (\url{https://www.youtube.com/watch?v=outcGtbnMuQ}) the human emotion capability of GPT-4 compared with the ArtGPT-4}
\label{Art2}
\end{table*}

\begin{table*}[htbp]
\centering
\begin{tabular}{|lp{0.8\textwidth}|}
\hline
\multicolumn{2}{|l|}{\textbf{ArtGPT-4 and image example compare}} \\
\hline
\textbf{Ures} & What’s funny about this image? \\
 & \includegraphics[width=6cm]{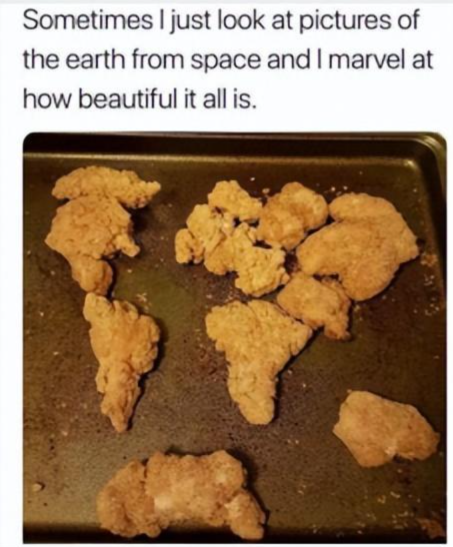} \\
\hline
\textbf{GPT-4} & This meme is a joke that combines two unrelated things: pictures of the earth from space and chicken nuggets.
The text of the meme suggests that the image below is a beautiful picture of the earth from space. However, the image is actually of chicken nuggets arranged to vaguely resemble a map of the world. The humor in this meme comes from the unexpected juxtaposition of the text and the image. The text sets up an expectation of a majestic image of the earth, but the image is actually something mundane and silly. \\
\hline
\textbf{ArtGPT-4} & The humor in this image comes from the idea of using food to represent a map of the world and the absurdity of using such a small and seemingly insignificant object to represent something as large and complex as the Earth. The fact that the image is presented as a meme on social media further adds to the humor by making it seem like a silly and unexpected thing to do.\\
\hline
\end{tabular}
\caption{In the example artical~\citep{openai2023gpt4} the human emotion capability of GPT-4 compared with the ArtGPT-4}
\label{Art3}
\end{table*}

As shown in Table~\ref{Art2} and Table~\ref{Art3}, the ArtGPT-4 performs remarkably well in understanding human emotions. It even surpasses GPT-4 in certain aspects, such as its ability to accurately capture the squirrel's expression. It provides a rich vocabulary and expressive language when describing the image, such as the use of "cute" and "playful", which effectively convey the image's sense of humor and charm. Moreover, it is able to describe facial expressions and actions depicted in the image, which is an important aspect of understanding human emotions.

\subsection{Abnormal Image Understanding Skills}

\begin{table*}[htbp]
\centering
\begin{tabular}{|lp{0.8\textwidth}|}
\hline
\multicolumn{2}{|l|}{\textbf{ArtGPT-4 and image example compare}} \\
\hline
\textbf{Ures} & What is unusual about this image? \\
 & \includegraphics[width=6cm]{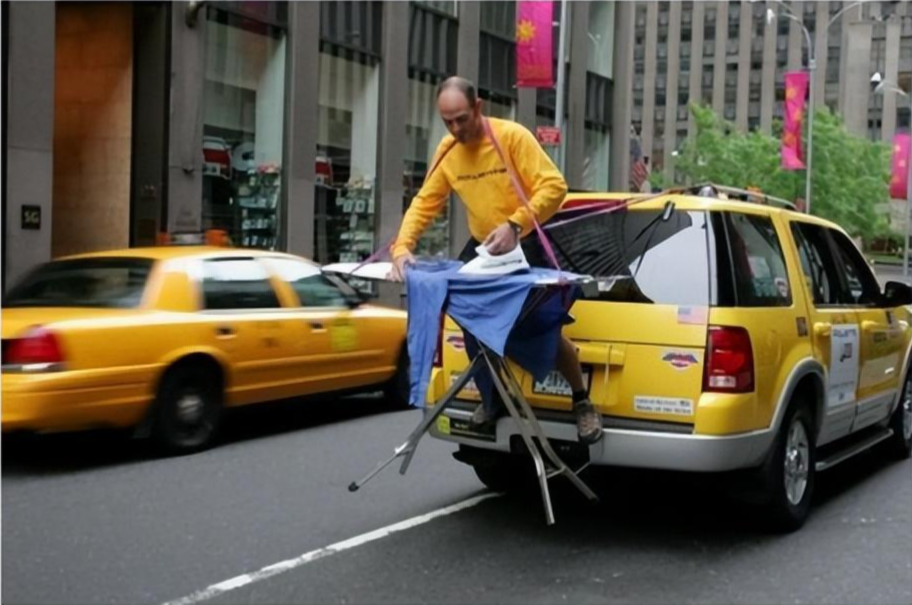} \\
\hline
\textbf{GPT-4} & The unusual thing about this image is that a man is ironing clothes on an ironing
board attached to the roof of a moving taxi. \\
\hline
\textbf{ArtGPT-4} & It is unusual because it shows a man standing on the back of a taxi holding an iron in his hand and ironing clothes. This is not a common sight in most cities.\\
\hline
\end{tabular}
\caption{In the example artical~\citep{openai2023gpt4} the abnormal image capability of GPT-4 compared with the ArtGPT-4}
\label{Art4}
\end{table*}

As shown in Table~\ref{Art4}, ArtGPT-4 can depict the abnormalities of the image as well as GPT-4.

\begin{table*}[htbp]
\centering
\begin{tabular}{|lp{0.8\textwidth}|}
\hline
\multicolumn{2}{|l|}{\textbf{ArtGPT-4 and image example compare}} \\
\hline
\textbf{Ures} & What is unusual about this image? \\
 & \includegraphics[width=10cm]{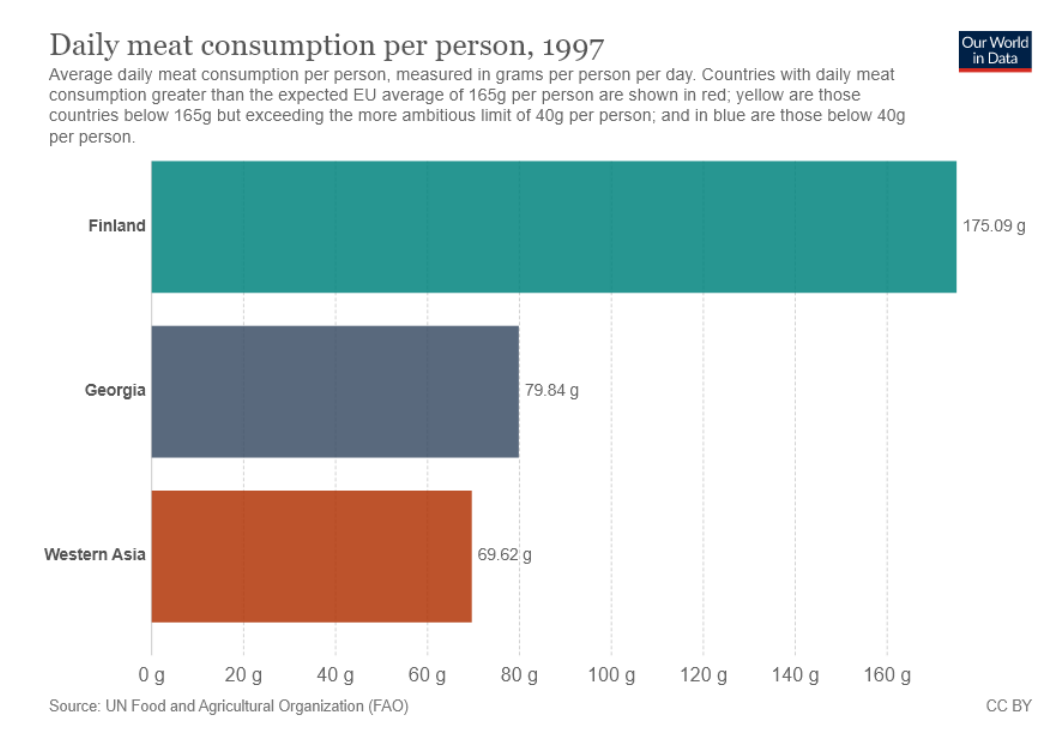} \\
\hline
\textbf{GPT-4} & To find the sum of average daily meat consumption for Georgia and Western Asia,
we will follow these steps:

1. Identify the average daily meat consumption for Georgia.

2. Identify the average daily meat consumption for Western Asia.

3. Add the values from steps 1 and 2.

Step 1: From the provided chart, we see that the average daily meat consumption for Georgia is 79.84 grams per person per day.

Step 2: From the chart, we see that the average daily meat consumption for
Western Asia is 69.62 grams per person per day.

Step 3: Add the values from steps 1 and 2.
79.84 g (Georgia) + 69.62 g (Western Asia) = 149.46 g

So, the sum of average daily meat consumption for Georgia and Western
Asia is 149.46 grams per person per day. \\
\hline
\textbf{ArtGPT-4} & To find the sum of average daily meat consumption for Georgia and Western Asia, we need to look at the data provided in the graph.

Georgia’s average daily meat consumption per person in 1997 was 79.84 grams.

Western Asia’s average daily meat consumption per person in 1997 was 69.62 grams.

The sum of the average daily meat consumption per person for Georgia and Western Asia in 1997 is 79.84 + 69.62 = 149.46 grams.

Therefore, the sum of average daily meat consumption per person for Georgia and Western Asia in 1997 is 149.46 grams.\\
\hline
\end{tabular}
\caption{In the example artical~\citep{openai2023gpt4} the analysis data capability of GPT-4 compared with the ArtGPT-4}
\label{Art5}
\end{table*}

\subsection{Data Analysis Capability}

ArtGPT-4, exhibits an exceptional ability to comprehend mathematical problems and solve them methodically, similar to human cognition. Moreover, its performance is on par with the renowned GPT-4. As shown in Table~\ref{Art5}, the model accurately identifies and extracts relevant information from a graph to derive the sum of average daily meat consumption for Georgia and Western Asia in 1997. Its proficiency in solving such problems showcases the model's potential to contribute to various fields, including data analysis and decision-making.

\end{document}